\makeatletter
\@namedef{ver@everyshi.sty}{}
\makeatother

\documentclass[final]{cvpr}

\usepackage{times}
\usepackage{epsfig}
\usepackage{graphicx}
\usepackage{amsmath}
\usepackage{amssymb}
\usepackage{relsize}
\usepackage{color}
\usepackage[usenames,dvipsnames]{xcolor}
\usepackage{booktabs}
\usepackage{amsmath}
\usepackage{subcaption}
\usepackage{tikz}
\usepackage{multirow}

\usepackage[pagebackref=true,breaklinks=true,letterpaper=true,colorlinks,bookmarks=false]{hyperref}

\definecolor{ubpubColor}{rgb}{0.43, 0.5, 0.5}
\definecolor{backboneColor}{rgb}{0.423, 0.325, 0.365}
\definecolor{fpnColor}{rgb}{0.255, 0.498, 0.416}

\newcommand{\PAR}[1]{\vskip4pt \noindent {\bf #1~}}

\newcommand{\embd}[1]{$\mathlarger{\mathlarger{\varepsilon}}$}

\newcommand*{\metric}{{\it LSTQ}\@\xspace} 

\newcommand*{\thing}{{\it thing}\@\xspace}
\newcommand*{\stuff}{{\it stuff}\@\xspace} 

\newcommand*{\thingset}{{\it things}}

\newcommand*{\gt}{{\it gt}} 
\newcommand*{\pr}{{\it pr}} 

\newcommand*{\gtc}{{\it {gt}_{\text{agn}}}} 
\newcommand*{\prc}{{\it {pr}_{\text{agn}}}} 

\newcommand*{\gti}{{\it {gt}_{id}}} 
\newcommand*{\pri}{{\it {pr}_{id}}} 

\newcommand*{\sfd}{\metric} 
\newcommand*{\sassoc}{\it S_{assoc}} 
\newcommand*{\sseg}{\it S_{cls}} 

\newcommand*{\tp}{\text{TP}_c} 
\newcommand*{\fn}{\text{FN}_c} 
\newcommand*{\fp}{\text{FP}_c}

\newcommand*{\tpa}{{\it TPA}} 
\newcommand*{\fna}{\it {FNA}} 
\newcommand*{\fpa}{\it {FPA}} 

\newcommand*{\iou}{{\it IoU}} 
\newcommand*{\tubesset}{\mathcal{T}}
\newcommand*{\pt}{\mathbf{p}}

\def\cvprPaperID{3346} %
\def\confYear{CVPR 2021}

\begin{document}

\title{4D Panoptic LiDAR Segmentation}

\author{Mehmet Ayg\"un$^1$\footnotemark[1]
\quad
Aljo\u sa O\u sep$^1$\footnotemark[1]
\quad
Mark Weber$^1$
\quad
Maxim Maximov$^1$\\
\quad
Cyrill Stachniss$^2$
\quad
Jens Behley$^2$
\quad
Laura Leal-Taix\'{e}$^1$\\
$^1$Technical University of Munich, Germany \quad $^2$University of Bonn, Germany\\
$^1${\tt\small \{mehmet.ayguen, aljosa.osep, leal.taixe, mark-cs.weber, maxim.maximov\}@tum.de} \\
\quad
$^2${\tt\small \{firstname.lastname\}@igg.uni-bonn.de}
}

\twocolumn[{%
\renewcommand\twocolumn[1][]{#1}%
\maketitle
\begin{center}
\vspace{-0.55cm}
  \includegraphics[width=0.965\linewidth]{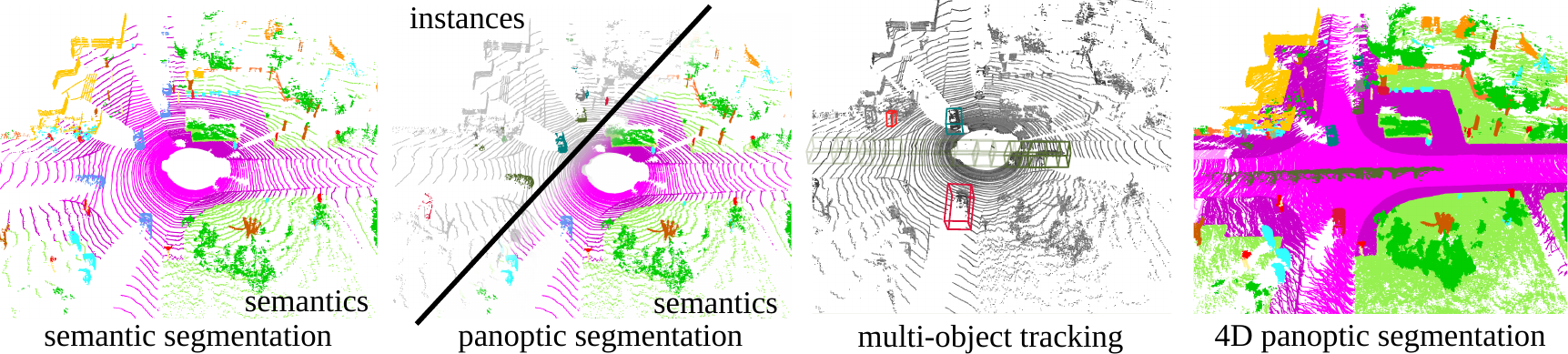}
    \captionof{figure}{Types of LiDAR-based scene understanding. Semantic and panoptic segmentation assign  semantic classes and determine instances in 3D space. Multi-object tracking encompasses 3D object detection in space, followed by association over time. 4D panoptic LiDAR segmentation jointly tackles semantic and instance segmentation in 3D space over time.}
    \label{fig:teaser}
    \vspace{0.1cm}
\end{center}%
}]

\begin{abstract}
\vspace{-0.2cm}
Temporal semantic scene understanding is critical for self-driving cars or robots operating in dynamic environments. In this paper, we propose \textit{4D panoptic LiDAR segmentation} to assign a semantic class and a temporally-consistent instance ID to a sequence of 3D points. To this end, we present an approach and a point-centric evaluation metric.
Our approach determines a semantic class for every point while modeling object instances as probability distributions in the 4D spatio-temporal domain. We process multiple point clouds in parallel and resolve point-to-instance associations, effectively alleviating the need for explicit temporal data association. Inspired by recent advances in benchmarking of multi-object tracking, we propose to adopt a new evaluation metric that separates the semantic and point-to-instance association aspects of the task. With this work, we aim at paving the road for future developments of temporal LiDAR panoptic perception.
\vspace{-0.5cm}
\end{abstract}

\makeatletter{\renewcommand*{\@makefnmark}{}

\footnotetext{* Authors contributed equally.}\makeatother}
\section{Introduction}

Spatio-temporal interpretation of raw sensory data is important for autonomous vehicles to understand how to interact with the environment and
perceive how trajectories of moving agents evolve in 3D space and time.

In the past, different aspects of dynamic scene understanding such as 
semantic segmentation~\cite{Everingham10IJCV, dai17cvpr, Milioto19IROS,Thomas19ICCV, zhang20CVPR, tang20eccv}, object detection~\cite{Felzenszwalb08CVPR, Ren15NIPS, Lang19CVPR, Shi19CVPR, shi20cvpr, shi20pami}, instance segmentation~\cite{He17ICCV}, and multi-object tracking~\cite{Leibe08TPAMI, Bergmann19ICCV, Osep17ICRA, Braso20CVPR, Weng20iros, qi20cvpr, poschmann20arxiv} have been tackled independently. %
The developments in these fields were largely fueled by the rapid progress in deep learning-based image~\cite{Krizhevsky12NIPS} and point-set representation learning~\cite{Qi17CVPR_pointnet, Qi17NIPS, Thomas19ICCV}, together with contributions of large-scale datasets, benchmarks, and unified evaluation metrics~\cite{Lin14ECCV, Everingham10IJCV, Geiger12CVPR, dendorfer20ijcv, Cordts16CVPR, Voigtlaender19CVPR, Gupta19CVPR, Behley19ICCV, dai17cvpr, Caesar20CVPR, sun20CVPR}. 
In the pursuit of image-based holistic scene understanding, recent community efforts have been moving towards convergence of tasks, such as multi-object tracking (MOT) and segmentation~\cite{Voigtlaender19CVPR, Yang19ICCV}, and semantic and instance segmentation, \ie, panoptic segmentation~\cite{Kirillov19CVPR}.
Recently, panoptic segmentation was extended to the video domain~\cite{kim20cvpr}. Here, the dataset, task formalization, and evaluation metrics focused on interpreting short and sparsely labeled video snippets in 3D (2D image+time) in an offline setting. Autonomous vehicles, however, need to continuously interpret sensory data and localize objects in a 4D continuum. %

Tackling sequence-level LiDAR panoptic segmentation is a challenging problem, since state-of-the-art methods~\cite{Thomas19ICCV} usually need to downsample even single-scan point clouds to satisfy the memory constraints. %
Therefore, the common approach in (3D) multi-object tracking is detecting objects in individual scans, followed by temporal association~\cite{Frossard18ICRA, Weng20iros, weng20CVPR}, often guided by a hand-crafted motion model. 
In this paper, we take a substantially different approach, inspired by the unified space-time treatment philosophy. We form overlapping 4D volumes of scans (see Fig.~\ref{fig:teaser}) and, in parallel, assign to 4D points a semantic interpretation while grouping object instances jointly in 4D space-time. 

Importantly, these 4D volumes can be processed in a single network pass, and the temporal association is resolved implicitly via clustering. This way, we retain inference efficiency while resolving long-term association between overlapping volumes based on the point overlap, alleviating the need for explicit data association. 

For the evaluation, we introduce a point-centric higher-order tracking metric, inspired by recent metrics for multi-object tracking~\cite{luiten20ijcv} and concurrent work on video panoptic segmentation~\cite{weber2021step} which differ from the available metrics~\cite{Kirillov19CVPR, Bernardin08JIVP} that overemphasize the recognition part of the tasks. Our metric consist of two intuitive terms, one measuring the semantic aspect and second the spatio-temporal association of the task. 
Together with the recently proposed SemanticKITTI~\cite{Behley19ICCV, Behley20arxiv} dataset, this gives us a test bed to analyze our method and compare it with existing LiDAR semantic/instance segmentation~\cite{Lang19CVPR, Thomas19ICCV, Weng20iros, milioto2020iros} approaches, adapted to the sequence-level domain.

In summary, our \textbf{contributions} are: (i) we propose a unified space-time perspective to the task of 4D LiDAR panoptic segmentation, and pose detection/segmentation/tracking jointly as point clustering %
which can effectively leverage the sequential nature of the data and process several LiDAR scans while maintaining memory efficiency; 
(ii) we adopt a point-centric evaluation protocol that fairly weights semantic and association aspects of this task and summarizes the final performance with a single number;
(iii) we establish a test bed for this task, which we use to thoroughly analyze our model's performance and the existing LiDAR panoptic segmentation methods used in conjunction with a tracking-by-detection mechanism. Our code, experimental data\footnote{\url{https://github.com/mehmetaygun/4d-pls}} and benchmark\footnote{\url{http://bit.ly/4d-panoptic-benchmark}} are publicly available.

\section{Related Work} 

Our work is related to tasks covering different aspects of scene perception, such as semantic segmentation, object detection/segmentation, and tracking. In the following, we review related methods and tasks. 

\PAR{Datasets and Metrics.}
The growing interest in autonomous vehicles has sparked interest in scene perception using LiDAR sensors.
Here the progress has been fueled by advances in deep learning on point sets~\cite{Qi17CVPR_pointnet, Qi17NIPS, hua18cvpr, komarichev19cvpr, lan19cvpr, Wu18ICRA, Wu19ICRA, Tatarchenko18CVPR, Thomas19ICCV, milioto2020iros} and datasets with standardized benchmarks for 3D semantic/instance segmentation~\cite{Geiger12CVPR, Behley19ICCV} and 3D object detection and multi-object tracking~\cite{Caesar20CVPR, sun20CVPR}. This confirms the importance of advancing both spatial and temporal aspects of mobile robot perception. 
Our proposed task formulation and evaluation metric is the first that unifies both aspects to the best of our knowledge.  

Recent community efforts in the field of image-based perception have been moving towards the convergence of different tasks. 
For instance, Kirillov \etal~\cite{Kirillov19CVPR} proposed to unify semantic and instance segmentation, which they termed panoptic segmentation, together with an evaluation metric, the panoptic quality (PQ). 
Others proposed to tackle multi-object tracking and instance segmentation~(MOTS) in videos jointly~\cite{Voigtlaender19CVPR,Yang19ICCV}.
Moreover, \cite{kim20cvpr} recently extended panoptic segmentation to videos -- however, the dataset and the evaluation metrics focus on interpreting short and sparse video snippets offline.  
This is reflected in the evaluation metric, which is essentially PQ evaluated based on the 3D IoU~\cite{Yang19ICCV} and averaged over temporal windows of varying sizes to compensate that the difficulty of the task depends on the sequence length. 
This setting is not suitable for autonomous vehicles that need to interpret raw sensor data continuously.  
Hurtado \etal ~\cite{hurtado20arxiv} proposes to combine ideas from MOTA~\cite{Bernardin08JIVP} and PQ~\cite{Kirillov19CVPR} by adding a penalty related to ID switches to the PQ. Nonetheless, both PQ and MOTA were criticized~\cite{porzi19cvpr,luiten20ijcv}, and the proposed evaluation inherits all of their well-known issues. 

In this paper, we propose a different approach and bring ideas recently introduced in the context of benchmarking vision-based multi-object tracking~\cite{luiten20ijcv} to the domain of sequential LiDAR semantic and instance segmentation. 
Together with the metric, we also propose an approach that operates directly on spatio-temporal point clouds providing object instances in space and time.

\PAR{Point Cloud Segmentation. } 
Semantic segmentation or point-wise classification of point clouds is a well-known research topic~\cite{anguelov05cvpr}. Traditionally, it was solved using feature extractors in combination with traditional classifiers \cite{agrawal09icra} and conditional random fields to enforce label consistency of neighboring points~\cite{triebel06icra, munoz09cvpr, xiong11icra}.
Availability of large-scale datasets, such as S3DIS~\cite{armeni16cvpr}, Semantic3D~\cite{hackel2017isprs}, and recently SemanticKITTI~\cite{Behley19ICCV}, made it possible to also investigate end-to-end pipelines~\cite{landrieu18cvpr, Milioto19IROS, Thomas19ICCV, hu20cvpr, zhang20CVPR, shi20CVPRSpSequenceNet, Qi17NIPS, Qi17CVPR_pointnet, milioto2020iros}. Similar to recent trends in RGB-D~\cite{jiang20cvpr_pointgroup, Engelmann20CVPR} and LiDAR segmentation~\cite{wong20corl}, our method performs bottom-up point grouping in a data-driven fashion. However, different from the aforementioned, we perform grouping in 3D space and time. We use the backbone by~\cite{Thomas19ICCV} that applies deformable point convolutions directly on the point clouds. In our case, this empirically performed better compared to backbones, specifically designed for point sequences~\cite{choy20194d, shi20CVPRSpSequenceNet}.

\PAR{Multi-Object Tracking and Segmentation.} 
The majority of vision-based MOT methods follow tracking-by-detection~\cite{Okuma04ECCV}. 
Here the idea is to first run a pre-trained object detector independently in each video frame and then associate detections across time. 
In the past, there was a strong focus on developing robust and, preferably, globally optimal methods for data association~\cite{Zhang08CVPR,Leal11ICCVW,Pirsiavash11CVPR,Brendel11CVPR,ButtCollins13CVPR}. 
Recent data-driven trends mainly focus on learning to associate detections~\cite{LealTaixe16CVPRW, Son17CVPR} or to regress targets~\cite{Bergmann19ICCV}, often in combination with end-to-end learning~\cite{peng19ICCV, xu19cvpr, Braso20CVPR}.  

In the realm of robot vision, it is critical to localize object trajectories in 3D space and time. Early methods localized monocular detections in 3D \eg, using  stereo~\cite{Leibe08TPAMI, Osep17ICRA, Ess09PAMI}, or performed tracking in a  category-agnostic manner by first performing bottom-up segmentation based on spatial proximity followed by point-segment association~\cite{Teichman12IJRR, Held14RSS}. 
Recently, LiDAR-based MOT became popular, thanks to the emergence of reliable 3D object detectors~\cite{Shi19CVPR, Lang19CVPR} and LiDAR-centric datasets~\cite{Caesar20CVPR, sun20CVPR}. Weng \etal~\cite{Weng20iros} demonstrated that simple methods based on linear assignment and constant-velocity motion models can perform surprisingly well when 3D detections are localized reliably. 
Our method departs from 3D object detection in the spatial domain, followed by the detection association in the temporal domain. Instead, we follow recent advances in image~\cite{Neven19CVPR, cheng20cvpr} and video instance segmentation~\cite{Athar20eccv}. 
We localize possible object instance centers within a 4D volume and associate points to estimated centers in a bottom-up manner, while a semantic branch assigns semantic classes to points.

\section{Method}

\label{sec:method}

In this paper, we propose a method and a metric for {4D Panoptic LiDAR Segmentation} task that tackles LiDAR {semantic segmentation} and {instance segmentation} jointly in the spatial and temporal domain.  
Given a sequence of LiDAR scans, the goal of this task is to predict for each 3D point (i) a semantic label for both \textit{stuff} and \textit{thing} classes, and (ii) a unique, identity-preserving object instance ID that should persist over the whole sequence.

\subsection{4D Panoptic LiDAR Segmentation: 4D-PLS}%

In this work, we take a different path compared to the tracking-by-detection paradigm to video-instance and video-panoptic segmentation~\cite{Voigtlaender19CVPR, Yang19ICCV, kim20cvpr, hurtado20arxiv}. We pose {4D panoptic segmentation} as two joint processes. The first one is responsible for point grouping in the 4D continuum using clustering, while the second assigns a semantic interpretation to each point.

\begin{figure}[t]
     \centering
        \includegraphics[width=\linewidth, trim=0 0 0 0, clip]{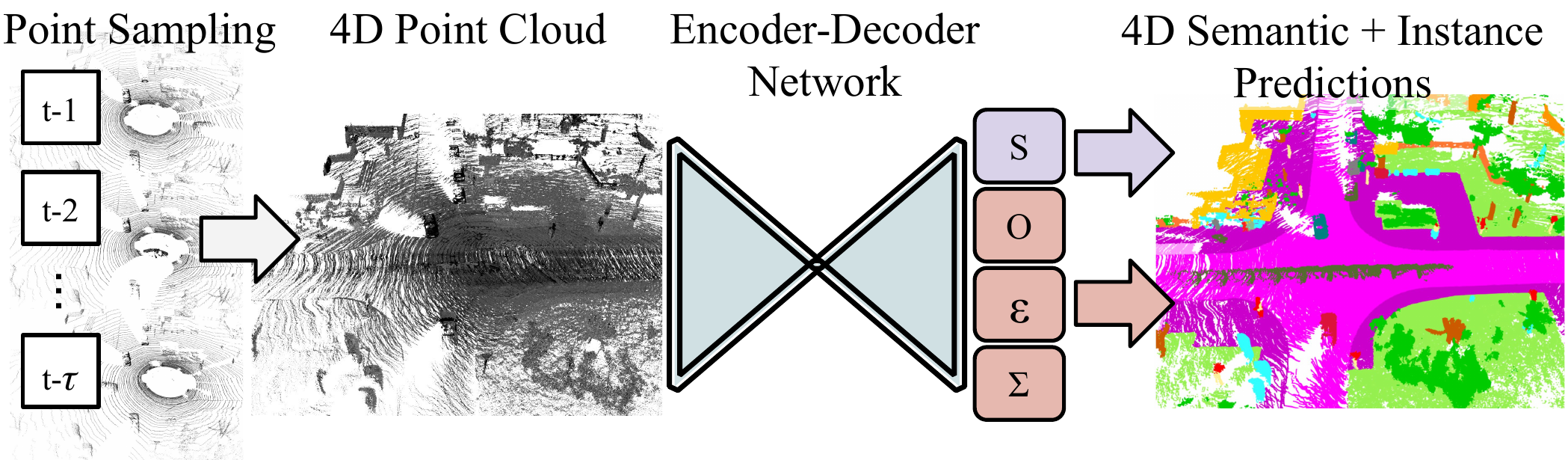}
    \vspace{-18pt}
    \caption{Visualization of our method. We sample points from past scans to form a 4D point cloud. Our encoder-decoder network estimates point objectness map ($O$), point variance map ($\Sigma$), and point embeddings ($\varepsilon$). We use these maps to assign points to their respective instances via density-based clustering in a 4D continuum. We obtain semantic interpretation from the semantic decoder ($S$).}
    \label{fig:main_fig}
\end{figure}

We provide an overview of our method in Fig.~\ref{fig:main_fig}. In a nutshell, we first form 4D point clouds from several consecutive LiDAR scans. 
In parallel, within a single network pass, we localize the most likely object centers (inspired by point-based tracking methods by~\cite{zhou20ECCV, cheng20cvpr}) in the sequence (objectness map $O$), assign semantic classes to points (semantic map $S$), and compute per-point embeddings (embedding map $\varepsilon$) and variances (variance map $\Sigma$).

The clustering can be performed efficiently by evaluating the probability of each 4D point belonging to a certain ``seed'' point, which is similarly performed in the context of images and video segmentation~\cite{Neven19CVPR, Athar20eccv}.
Finally, to associate 4D sub-volumes, we examine point intersections between overlapping point volumes.

\PAR{4D Volume Formation.}
During inference and training, we form overlapping 4D point cloud volumes in an online setting. 
In particular, for scan $t$ and temporal window size $\tau$, we align together point clouds within temporal window $\{\max(0, t-\tau), ..., t\}$ using ego-motion estimates provided by a SLAM approach~\cite{behley18rss}. 
Our experiments in Sec.~\ref{sec:ablation} reveal that processing multiple point clouds significantly improves spatial and temporal point association performance. However, due to the linear growth in memory requirements, stacking point clouds along the temporal dimension is prohibitively expensive. 
To overcome this issue, we build on the intuition that \textit{thing} classes are most critical for a stable temporal association, since these classes correspond to potentially moving objects. As we operate in an online setting, where the past scans have already been processed, we can sample points that belong to \textit{thing} classes or lie near to an object centers from earlier scans.

\PAR{Density-based Clustering.} 
We model object instances via Gaussian probability distributions. 
Given an estimate of the object center, \ie, clustering ``seed'' point, we can assign points to their respective instance by evaluating each point under the Gaussian pdf based on the point's embedding vectors. 
The estimated centers do not need to correspond to exact object centers but are merely used to initiate the clustering. Thus, our approach is in practice fairly robust to occlusions and cross-time view changes. We note that the Gaussian assumption is only valid for shorter temporal windows. 
In particular, given a point $\pt_i$ representing the instance center and its  embedding vector $e_{i}$, and a query point $p_j$ with its embedding vector $e_j$, we can evaluate the probability of point $p_j$ belonging to its center ``seed'' point $p_i$ as:
\begin{align}
\label{eq:gaussian}
    \hspace{-0.1cm}\hat{p}_{ij} &= \frac{1}{(2\pi)^{\frac{D}{2}}|\Sigma_{i}|^{\frac{1}{2}}}\exp{\hspace{-0.1cm}\left(\hspace{-0.1cm} -\frac{1}{2}(e_{i}-e_{j})^{\top}\Sigma_{i}^{-1}(e_{i}-e_{j}) \hspace{-0.1cm}\right)\hspace{-0.1cm}}
\end{align}
where $\Sigma_{i}$ is a diagonal matrix constructed using variance prediction $\sigma_{i}$ of point $p_{i}$. We concatenate coordinate values $(x, y, z, t)$ with the learned point embedding vectors to combine spatial and temporal coordinates with learned embeddings. 
We account for these additional dimensions during the training of the variance map.

\PAR{Network and Training.}
To perform such clustering, we need to identify most likely instance centers, \ie, ``seed'' points, in a 4D point cloud. We also need variance predictions for each point to evaluate probability scores during clustering, and a posterior over all semantic classes.  

We estimate all these quantities using an encoder-decoder architecture that operates directly on the 4D point cloud $P \in \mathbb{R}^{N \times 4}$. 
The encoder network is based on the KP-Conv~\cite{Thomas19ICCV} backbone that uses deformable point convolutions. 
The decoder predicts point-wise feature embeddings $\mathlarger{\mathlarger{\varepsilon}} \in \mathbb{R}^{N \times D}$ using consecutive point convolutions. 
On top of the encoder, we add an object centerness decoder in $\mathbb{R}^{N \times 1}$, point variance decoder in $\mathbb{R}^{N \times D}$, and semantic decoder in $\mathbb{R}^{N \times C}$. We train our network in an end-to-end manner and in online fashion.

To train the semantic decoder, we use cross-entropy classification loss $L_\text{class}$. 
As the semantic classes are highly imbalanced, we sample points to ensure that the probability of sampling a point from a certain class is roughly uniform.

To learn the point centerness and point variance, we use three different losses. First, we impose the mean squared error (MSE) loss to train the object centerness decoder. 
However, there will generally be no points near the actual object centers, unlike the image and video domain~\cite{Neven19CVPR,Athar20eccv}. 
Therefore, instead of predicting per-point centerness, we predict the proximity of the point to its instance center. We compute for each point $\pt_i$ its objectness $o_{i}$ as Euclidean distance between the point and its instance center, \ie, mean point of all instance points, normalized to $[0, 1]$. This objectness $o_{i}$ is then compared to the regressed objectness $\hat{o}_{i}$:
\begin{equation}
  \label{eq:obj}
 L_{\text{obj}} = \sum_{i=1}^{N} 
   \left(\hat{o}_{i} - o_{i}\right)^{2}, \;\;\;\; \hat{o}_{i}, o_{i} \in [0,1]
\end{equation}
Since we want the embeddings of instances to form clusters in the spatio-temporal domain, we introduce our instance loss. Given a 4D point cloud of $N$ points and $K$ instances, it is defined as:
\begin{equation}
  \label{eq:ins}
  L_{\text{ins}} = \sum_{j=1}^{K}\sum_{i=1}^{N}
   \left(\hat{p}_{ij} - p_{ij}\right)^{2}, 
   p_{ij} = 
   \begin{cases}
      1, & \text{if}\ \pt_{i} \in I_{j} \\
      0, & \text{otherwise}
    \end{cases}
\end{equation}
where $\hat{p}_{ij}$ is evaluated under the Gaussian pdf (Eq.~\ref{eq:gaussian}) with points embedding $e_{i}$ as well as instance embedding and variance $e_{j}$ and $\sigma_{j}$. 
In addition, we employ variance smoothness loss $L_\text{var}$, similar to~\cite{Athar20eccv, Neven19CVPR} for training the variance decoder. In summary, we use four different losses to train our network in end-to-end manner: $L = L_\text{class}+ L_\text{obj} + L_\text{ins} + L_\text{var}$.

\PAR{Inference.} 
We resolve point-to-instance associations in two stages, first within a processed 4D volume, and then across volumes. 
First, based on the point cloud centerness map, we select the point $\pt_{i}$, which has the highest objectness score.
Then, we evaluate probabilities $p_{ij}$ for all candidate points and assign them to the cluster in case $p_{ij} > 0.5$. The assigned points are then removed from the candidate pool. We repeat these steps until the next highest objectness score is below a certain threshold. To transfer identities across processed 4D volumes, we perform cross-volume association greedily based on the overlap score, taking all scans into account. When the overlap is below a threshold, we assign a new id.

\subsection{Measuring Performance}\label{sec:metrics}

The central question when proposing a novel task and benchmark is how to evaluate and compare different methods.
Preferably, we would like to summarize performance with a single number to rank the methods while retaining the capability of looking at different aspects of the task. 

\subsubsection{Existing Evaluation Measures}\label{sec:metrics}

To motivate our approach to evaluation, we first briefly discuss established metrics for image-based panoptic segmentation (PQ~\cite{Kirillov19CVPR}) and multi-object tracking and segmentation (MOTSA/MOTSP~\cite{Bernardin08JIVP, Voigtlaender19CVPR}). Then, we discuss two recently proposed extensions of PQ to the temporal domain and argue why we do not promote their adaptation for the task of 4D LiDAR panoptic segmentation. 

\PAR{Segment-centric Evaluation.} %
PQ and MOTSA/MOTSP are instance-centric evaluation metrics. Both first determine a unique matching between sets of ground-truth objects and model predictions for each frame individually to determine true positives (TPs), false positives (FPs), and false negatives (FNs). 
Both metrics provide measures for the segmentation and recognition aspects of the task. 
The segmentation quality (SQ) term of PQ and MOTSP integrates IoUs over the set TPs and normalizes it by the size of the TP set. 
The recognition quality (RQ) term of PQ is expressed as the $F_1$ score. 
Similarly, MOTSA combines detection errors (FNs and FPs) with ID switch (IDSW) penalty in a single term. IDSW occurs when a track is lost, and the tracker assigns a new identity to a tracked object. This is the only term that takes the temporal aspect of the task into account. 

A criticism of PQ is that it over-emphasizes the importance of very small segments and \stuff classes can be difficult to match~\cite{porzi19cvpr}. 
MOTSA overemphasizes the detection compared to the association aspect and it is nonintuitive, since the score can be negative and is unbounded, as can be seen in Sec.~\ref{sec:experiments}. 
Furthermore, the influence of the ID switches on the final score depends on the frame rate, and MOTSA does not reward trackers that recover from incorrect associations. 
Importantly, both metrics are sensitive to the selection of the matching threshold.  
Thus, instances that slightly miss this threshold will cause both a FN and a FP. 
This is not the case for pixel or point centric metrics used for evaluating semantic segmentation. The standard mean IoU~(mIoU) metric~\cite{Everingham10IJCV} computes sets of TPs, FPs and FNs pixel (or point) basis, effectively bypassing the segment matching.

\PAR{PQ Extensions.} %
Recent work~\cite{kim20cvpr} proposes video panoptic quality, a variant of PQ for the sequential domain. Different from PQ, gt-to-prediction mapping is established based on the sequential IoU matching criterion~\cite{Yang19ICCV}.%
As objects are not present throughout the clip and the difficulty of the task critically depends on the length of the temporal window, the final metric is averaged over varying temporal window sizes. This is suitable for the setting defined in~\cite{kim20cvpr}, where the task is to evaluate short, sparsely labeled video snippets. However, this approach does not  scale to real-world sequences of arbitrary length. Another extension to PQ, panoptic tracking quality (PTQ)~\cite{hurtado20arxiv} combines MOTA and PQ by adding an ID penalty to the PQ measure. This approach inherits issues from both PQ and MOTSA metrics.

\subsubsection{LiDAR Segmentation and Tracking Quality}
\label{sec:our_metric}

\begin{figure*}[t]
     \centering
        \vspace{-0.0cm}
        \parbox{0.49\textwidth}{\hspace{3cm}\vspace{-0.25cm}Semantic Segmentation}
        \parbox{0.49\textwidth}{\hspace{3cm}\vspace{-0.25cm}Instance Segmentation}
        \rotatebox{90}{\resizebox{1.5cm}{!}{\parbox{2cm}{\center 2-scan prediction}}}
        \includegraphics[width=0.40\textwidth, trim=0 55 0 0, clip]{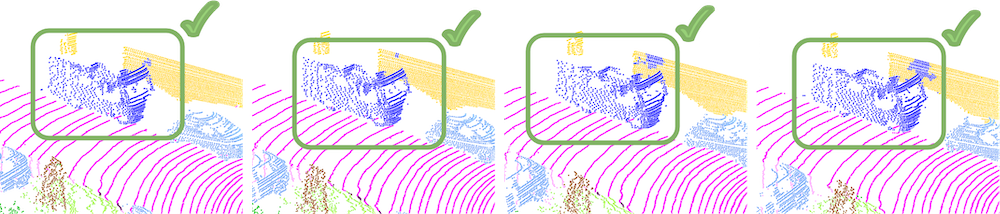}
        \begin{tikzpicture}[node distance=2cm]
        \draw (0,0) -- (0,1);
        \end{tikzpicture}
        \includegraphics[width=0.40\textwidth, trim=0 0 0 55, clip]{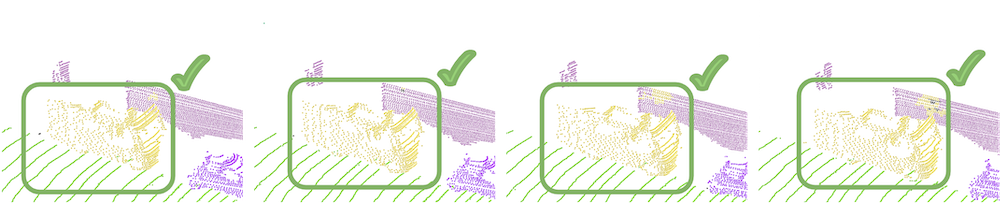}
        \rotatebox{90}{\resizebox{1.5cm}{!}{\parbox{2cm}{\center  4-scan prediction}}}
        \includegraphics[width=0.40\textwidth, trim=0 55 0 0, clip]{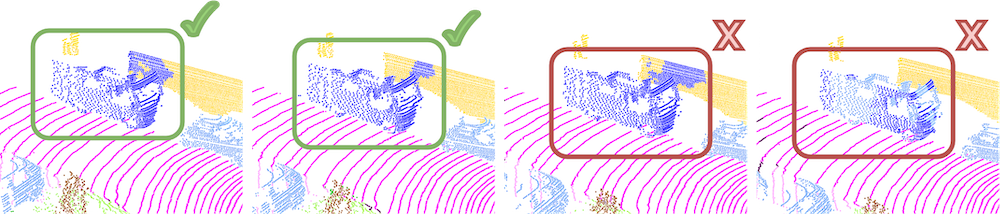}
        \begin{tikzpicture}[node distance=2cm]
        \draw (0,0) -- (0,1);
        \end{tikzpicture}
        \includegraphics[width=0.40\textwidth, trim=0 0 0 55, clip]{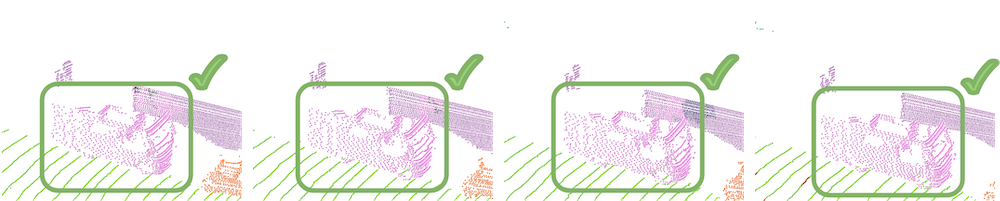}
        \begin{tikzpicture}[node distance=2cm]
        \node (A) at (0.25, 0.0) {};
        \node (B) at (15.5, 0.0) {};
        \draw[->, to path={-- (\tikztotarget)}](A) edge (B);
        \node[text width=1cm] at (16.0,0.0){time};
        \end{tikzpicture}
    \vspace{-6pt}
    \caption{Predictions from 2 and 4 scan versions, $\text{MOTSA=}1.0/0.0, S_{assoc=}0.96/0.93, S_{cls=}0.86/0.75$. While both models track the instance correctly, due to slight difference in semantic segmentation predictions, MOTSA scores differ drastically.}
    \label{fig:aqvsmotsa_main}
    \vspace{0.1cm}
\end{figure*}
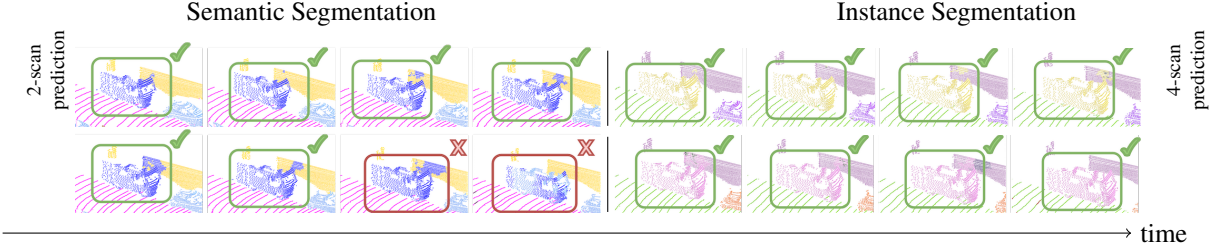

In the following, we assume a sequence of 3D point clouds of length $l$, sampled at discrete time-steps: $\Omega = \{(\pt, n) \in \mathbb{R}^3 \times \mathbb{N} | n < l \}$. 
We define the ground-truth assignment function as $\gt (\pt, n) \to (c, id)$ and a prediction function as $\pr (\pt, n) \to (c, id)$, that map each 4D tuple, consisting of a point $\pt$ and a timestamp $n$, to a certain class $c$ and identity $id$. 
In the following, we devise an evaluation metric that, for each pair $(\pt, n)$, evaluates whether (i) it was assigned to a correct class, and (ii) for the \thing classes, whether it was assigned to the correct object instance. 
Inspired by the recently introduced Higher Order Tracking Accuracy (HOTA)~\cite{luiten20ijcv}, proposed in the context of MOT, and concurrent work on video panoptic segmentation proposing the Segmentation and Tracking Quality (STQ)~\cite{weber2021step}, our \sfd (LiDAR Segmentation and Tracking Quality) consists of two terms, the classification score $\sseg$ and the association score $\sassoc$. 

We adopt a fundamentally different evaluation philosophy compared to other metrics~\cite{luiten20ijcv, Kirillov19CVPR, kim20cvpr, hurtado20arxiv}. In particular, we drop the concept of the frame-level ``detection`` and do not match segments between ground-truth and prediction. 
Our association score measures point-to-instance association quality in a unified way -- in space \textit{and} time at point level, which is more natural for segmentation tasks.

\PAR{Classification Score.}
For the classification score, we first define instance-agnostic ground-truth and predictions sets:
\begin{align*}
    \gtc (c) &= \{(\pt, n)\mid \text{gt}(\pt, n) = (c, *) \}, \\
    \prc (c) &= \{(\pt, n)\mid \text{pr}(\pt, n) = (c, *) \},
\end{align*}
representing the ground truth and predicted points that belong to class $c$ irrespective of their assigned ids. Then, the TP, FP, FN sets are computed as in semantic segmentation evaluation with respect to gt class $c$ and predicted class $c'$:
\begin{align*}
    \tp  &= |\prc(c) \cap \gtc(c)|,\\
    \fp  &= |\prc (c) - \prc (c) \cap \gtc(c)|,\\
    \fn  &= |\gtc (c) - \prc (c) \cap \gtc(c)|.
\end{align*}
The classification score then simply boils down to intersection-over-union (IoU) over these sets, which is the standard approach for evaluating semantic segmentation (however, this is different from segment-centric PQ, where points contribute to the $\tp$ term only if the segment that they belong is matched). We follow the standard procedure and average over the classes: 
\begin{align*}
    \sseg = \frac{1}{|\mathcal{C}|} \sum_{c=1}^{\mathcal{C}}  \frac{|\tp|}{|\tp| + |\fn| + |\fp|}  
    = \frac{1}{C} \sum_{c=1}^{\mathcal{C}} \text{\iou}(c).
\end{align*}

\PAR{Association Score.} 
To evaluate the association score, we introduce the following class-agnostic predictions and ground-truth for the \thing classes:
\begin{align*}
    \gti (id) &= \{(\pt, n)\mid gt(\pt, n) = (c, id), c \in \thingset \}, \\
    \pri (id) &= \{(\pt, n)\mid pr(\pt, n) = (c, id), c \in \thingset \}.
\end{align*}
We define the true positive association (\tpa) set between a ground-truth object $t$ with identity $id$ and prediction $s$, that was assigned identity $id'$. 
This gives us a set of points with mutually consistent identities $id$ and $id'$: %
\begin{equation}
    \label{eq:tpa}
   \tpa (id, id') = |\pri (id') \cap \gti(id)|.
\end{equation}
Similarly, we define the set of false positive associations: 
\begin{equation}
    \label{eq:fpa}
   \text{\fpa} (id, id') = |\pri (id') - \pri (id') \cap \gti(id)|.
\end{equation}
Intuitively, this set contains predicted point assignments with identity $id'$, that were assigned a different ground-truth identity ($\neq id$), or were not assigned to a valid object instance. 
Finally, the set of false negative assignments: 
\begin{equation}
    \label{eq:fna}
  \text{\fna} (id, id') = |\gti (id) - \pri (id') \cap \gti(id)|
\end{equation}
contains ground-truth points with identity $id$ that were assigned an identity, different to $id'$, or were missed. 
We note that the concept of TPA, FPA and FNA was first introduced in the context of MOT evaluation for measuring the quality of temporal detection association. 
Therefore, to establish these sets, a bijective mapping between gt and pred needs to be established (as in the case of ~\cite{Bernardin08JIVP}). However, in \sfd, these sets are established with respect to each 4D point, treating association in space and time in a unified manner. 
    
Once we have quantified these sets, we can evaluate how well a predicted segment $s$ \textit{agrees} with ground-truth segment $t$. 
Because a ground truth segment $t$ may be explained by multiple different predictions, we sum contributions of all pairs with non-zero overlap: 
\begin{align}
\label{eq:metric}
\hspace{-0.1cm}\sassoc&=\frac{1}{|\tubesset|} \sum_{t \in \tubesset} \frac{1}{|\gti(t)|}\hspace{-0.1cm}\sum_{\substack{ s \in S \\ s \cap t \neq 0 }}\hspace{-0.1cm}\tpa(s, t)\iou(s, t),
\end{align}
where the $\iou$ term is evaluated using the $\tpa, \fna$ and $\fpa$ sets (Eq.~\ref{eq:tpa},~\ref{eq:fna},~\ref{eq:fpa}). In practice, we do not need to perform any point segment association, and even a prediction with a single common point will contribute to this term. 
We normalize these contributions by the tube volume, and we weigh each contribution by the volume of the TPA set. This weighting term ensures that instances with larger temporal spans have a higher contribution to the final score. 
Finally, our metric is computed as a geometric mean of the two terms: $\text{\sfd} = \sqrt{ \sseg \times \sassoc }$. 
The advantage over the arithmetic mean is that the final score will become zero if any of the two terms approach zero. This reflects our intuition that failing at either of two aspects of the task should yield a very low final score.  

\sfd tolerates different semantic predictions within a spatio-temporal segment by design (the $\iou$ term in Eq.~\ref{eq:metric} is evaluated in a class-agnostic manner). Following STQ~\cite{weber2021step}, we \textit{decouple semantic and association errors}, otherwise, \eg, a truck mistaken for a bus would be harshly penalized by the association term, even though it was tracked correctly. This behavior that dis-entangles association and classification errors is different from MOTSA/PTQ/VPQ where semantics and temporal association are entangled.

\section{Experimental Evaluation}

\label{sec:experiments}

In this section, we first evaluate different strategies for forming 4D point cloud volumes, assess the impact of processing multiple scans on the final performance, and discuss several possibilities for the embeddings used for point grouping. We compare our method to baselines for single-scan LiDAR panoptic segmentation~\cite{Behley20arxiv} and 4D panoptic LiDAR segmentation by extending existing methods. %

We use the SemanticKITTI~\cite{Behley19ICCV} LiDAR dataset to conduct our experiments. It contains $22$ sequences from KITTI odometry dataset~\cite{Geiger12CVPR} and provides point-wise semantic and temporally consistent instance annotations~\cite{Behley20arxiv}. We use the training/validation/test split by SemanticKITTI~\cite{Behley19ICCV,Behley20arxiv}.

\subsection{Ablation studies}
\label{sec:ablation}

We perform all ablations on the validation set and interpret results through the lens of \metric metric~(Sec.~\ref{sec:our_metric}).

\PAR{Point Propagation.}
As discussed in Sec.~\ref{sec:method}, we cannot simply stack point clouds temporarily due to the memory constraints. We build on the intuition that we can subsample a set of points from the past scans that are most beneficial for the end-task performance. As we are operating in an online setting, and the past scans have  been already processed, we can leverage past predictions.

In this experiment, we discuss different temporal point sampling strategies for varying temporal window sizes of $\tau = {2, 4, 6}$.
In the \textit{thing-propagation} strategy, we exclusively sample points which are assigned only to a thing class, as they represent only a small number of all point.  
In the \textit{importance sampling} strategy, we sample 10\% of points with a  probability proportional to the objectness. This way, we focus on points likely to represent \thing classes, while still allowing to propagate points belonging to \stuff classes, which can aid the semantic segmentation of the task.  
Similarly, the \textit{temporal decay} sampling uses the objectness score as a deciding factor, but we decay the number of sampled points based on the proximity to the current scan. 
Finally, the \textit{strided sampling} samples points with a stride of $2$ along the temporal dimension. 

As can be seen in Tab.~\ref{table:abl_point}, the \textit{importance sampling} strategy yields higher performance compared to sampling only \thing classes, at a slightly increased memory cost. 
As expected, this approach improves association quality and aids semantics as it also propagates points representing \stuff.

\begin{table}[t]
\centering
\resizebox{1.0\linewidth}{!}{
\begin{tabular}{lc|ccc|cc|c}
\toprule
Strategy & \# fr.          & \metric           & S$_\text{assoc}$       & S$_\text{cls}$       & IoU$^\text{St}$     & IoU$^\text{Th}$             & Mem.          \\ \midrule

Base 
&1 & 51.92 & 45.16   & 59.69     & 64.60     & 60.40       & 1x     \\
\hline
Thing-prop.
&2        & 59.20           & 58.71             & \textit{59.69}    & \textit{64.04}    & \textit{61.16}    & 1.05x  \\
&4        & \textit{60.94}  & \textit{63.95}    & 57.84             & 63.95             & 56.67             & 1.15x  \\
&6        & 58.88           & 61.13             & 56.71             & 63.86             & 53.97             & 1.25x  \\
\hline
Importance samp. 
&2        & 59.86           & 58.79             & \textit{60.95}    & 64.96             & \textit{63.06}      & 1.1x   \\
&4        & \textbf{62.74}  & \textbf{65.11}    & 60.46             & \textit{65.36}    & 61.26            & 1.3x   \\
&6        & 61.52           & 64.28             & 58.88             & 65.32             & 57.38             & 1.5x   \\
\hline
Temporal decay  
&2        & 59.86           & 58.79             & \textit{60.95}    & 64.96             & \textit{63.06}    & 1.1x   \\
&4        & \textit{62.14}  & \textit{64.05}    & 60.30             & 65.33             & 60.91             & 1.3x   \\
&6        & 61.03           & 63.34             & 58.81             & \textit{65.38}    & 57.11             & 1.5x   \\
\hline
Temporal stride

&2        & \textit{62.29}  & \textit{63.52}    & \textbf{61.08}     & 65.10            & \textbf{63.18}      & 1.1x   \\
&4        & 61.63           & 62.89             & 60.39             & \textbf{65.45}     & 60.39      & 1.3x   \\
&6        & 59.33           & 59.95             & 58.72             & 65.39             & 56.88      & 1.5x  \\
 \bottomrule
\end{tabular}
}
\vspace{-10pt}
\caption{Ablation study on point sampling strategies for building 4D point cloud volumes with respect to different temporal window sizes. }
\label{table:abl_point}

\end{table}

\begin{table}[t]
\footnotesize
\centering
\resizebox{1.0\linewidth}{!}{
\begin{tabular}{ll|ccc|cc}
\toprule
Mixing & \# Sc.         & \metric  & S$_\text{assoc}$       & S$_\text{cls}$       & IoU$^\text{St}$     & IoU$^\text{Th}$            \\ \midrule
$xyz$ & 2 & 51.65 & 43.77 & 60.95 & 64.96 & 63.06  \\
$xyzt$ & 2 & 51.95 & 44.30 & 60.93 & 64.80 & 63.15  \\
\hline
Emb. & 2 & 48.58& 43.63& 54.10& 59.54& 50.41 \\
\hline
Emb. + $xyz$ & 2 & 54.15& 56.63& 55.11& 61.14& 53.71\\
Emb. + $xyzt$ & 2 & 59.86 & 58.79 & 60.95 & 64.96 & 63.06 \\
\hline
\hline
$xyz$ & 4 & 54.29 & 48.77 & 60.44 & 65.29 & 61.32  \\
$xyzt$ & 4 & 54.46 & 49.55 & 59.87 & 64.80 & 61.15  \\
\hline
Emb. & 4 & 56.77& 60.63& 53.17& 58.00& 52.25 \\
\hline

Emb. + $xyz$ & 4 & 58.43& 63.90& 54.68& 61.12& 52.67 \\
Emb. + $xyzt$ & 4 & 62.74 & 65.11 & 60.46 & 65.36 & 61.26  \\
 \bottomrule
\end{tabular}
}
\vspace{-5pt}
\caption{Embedding design ablation.}
\label{table:abl_mixing_track}
\end{table}

Interestingly, even a temporal window of size $2$ drastically improves the performance compared to a single scan baseline, at negligible memory consumption ($1.1\times$). 
We observe the largest gains when the scans are temporally close: our 4-scan multi-scan baseline improves performance $51.92 \to 62.74$ in terms of \sfd. The association term benefits more from processing multiple scans %
compared to the segmentation term. %
This confirms that our model learns to exploits temporal cues very well.  
While temporal decaying does aid semantic or temporal aspect, introducing a temporal stride of $2$ yields the highest performance gains for semantic point classification. However, denser sampling in the temporal domain benefits association.
Hence, we focus on the importance sampling strategy with $\tau = 4$. 
In the supplementary, we highlight the performance with temporal window size $\tau = {1, 2, 3, 4, 6, 8}$. 
As can be seen, the association accuracy is increasing up to $\tau=4$ and then saturates, while classification accuracy saturates a $\tau=2$; however, it only decreases marginally.

\begin{table}[t]
\centering
{
\resizebox{1.0\linewidth}{!}{
\begin{tabular}{lcccc|c}
\toprule
Method                                     & PQ            & $PQ^\dagger$          & SQ            & RQ       & mIoU          \\ 
\midrule
RangeNet++~\cite{Milioto19IROS}  + PointPillars~\cite{Lang19CVPR}  & 37.1          & 45.9          & 75.9          & 47.0       & 52.4          \\ 
KPConv~\cite{Thomas19ICCV} + PointPillars~\cite{Lang19CVPR}      & 44.5          & 52.5          & 80.0          & 54.4                & 58.8          \\ 
Panoptic RangeNet~\cite{milioto2020iros}      & 38.0        & 47.0        & 76.5          & 48.2                & 50.9          \\

\hline
Our method (single scan)                  & \textbf{50.3} & \textbf{57.8} & \textbf{81.6} & \textbf{61.0}  & \textbf{61.3} \\ \bottomrule

\end{tabular}
}
}
\vspace{-10pt}
\caption{Single scan Panoptic Segmentation (test set).}
\label{table:pan_results}
\end{table}

\begin{table}[t]
\centering
\resizebox{1.0\linewidth}{!}{
\begin{tabular}{l|ccc|cc}
\toprule
Method          & \metric           & S$_\text{assoc}$       & S$_\text{cls}$      & IoU$^\text{St}$     & IoU$^\text{Th}$  \\ \midrule
RangeNet++\cite{Milioto19IROS} + PP + MOT  & 35.52 & 24.06 & 52.43 & 64.52 & 35.82 \\ 
KPConv~\cite{Thomas19ICCV} + PP + MOT  & 38.01 & 25.86 & 55.86 & 66.90 & 47.66 \\ 

\hline
\hline
RangeNet++\cite{Milioto19IROS} + PP + SFP  & 34.91 & 23.25 & 52.43 & 64.52 & 35.82  \\ 
KPConv~\cite{Thomas19ICCV} + PP + SFP & 38.53 & 26.58 & 55.86 & 66.90 & 47.66 \\ 
\hline
\hline
Our (single scan) + MOT  & 40.18 & 28.07 & 57.51 & 66.95 & 51.50 \\ 
Our (single scan) + SFP  & 43.88 & 33.48 & 57.51 & 66.95 & 51.50  \\ 

Ours (multi scan) & 56.89 & 56.36 & 57.43 & 66.86 & 51.64 \\

\bottomrule
\end{tabular}
}
\vspace{-7pt}
\caption{4D Panoptic (test set). MOT -- \textit{tracking-by-detection} by~\cite{Weng20iros}; SFP -- \textit{tracking-by-detection} via scene flow based propagation~\cite{mittal20cvpr}; PP -- PointPillars~\cite{Lang19CVPR}.}
\label{table:baselines4d}
\end{table}

\PAR{Embedding Design.} 
In this experiment, we study different point embeddings for clustering and show our findings in Tab.~\ref{table:abl_mixing_track}. 
We investigate the base performance of using only 3D spatial $(xyz)$ and 4D spatio-temporal point coordinates $(xyzt)$, and using only learned embeddings (Emb.). Next, we investigate the performance of the coordinate mixing formulation that combines learned embeddings with 3D spatial and 4D spatio-temporal coordinates. As can be seen, the variant in which we combine both yields the best results, not only in terms of $\sassoc$, but also $\sseg$. This shows that a well-designed embedding branch has a positive effect on learning the backbone features. 
Note that for the baseline that uses only spatio-temporal coordinates, we still use our fully trained network.

\subsection{Benchmark Results}

\PAR{Single-scan Prediction.}
First, we evaluate our method using single-scan LiDAR panoptic segmentation~\cite{Behley19ICCV, Behley20arxiv} to demonstrate the effectiveness of our network solely in the spatial domain. 
We use points from single scans during training and testing. We follow the standard evaluation protocol and compare to published and peer-reviewed methods.

\begin{table*}[ht!]
\centering
\resizebox{1.0\linewidth}{!}{
\begin{tabular}{l|ccc|cc|cc|ccc}
\toprule
Method          & \metric           & S$_\text{assoc}$       & S$_\text{cls}$      & IoU$^\text{St}$     & IoU$^\text{Th}$    & sPTQ           & PTQ            & sMOTSA         & MOTSA   \\ \midrule
RangeNet++\cite{Milioto19IROS} + PP + MOT  & 43.76 & 36.28 & 52.78 & 60.49 & 42.17  & 34.58  & 33.83  & {-7.88} & {-4.57} \\ 
KPConv~\cite{Thomas19ICCV} + PP + MOT   & {46.27} & {37.58} & {56.97} & {64.21} & {54.13}  & {39.13}  & {38.11}  & -6.16 & -2.41 \\ 

\hline
\hline
RangeNet++\cite{Milioto19IROS} + PP + SFP  & 43.38 & 35.66 & 52.78 & 60.49 & 42.17  & 35.83  & 35.46  & {-3.13} & {-0.01}  \\ 
KPConv~\cite{Thomas19ICCV} + PP + SFP & {45.95} & {37.07} & {56.97} & {64.21} & {54.13}  & {41.44}  & {41.05}  & 2.83 & 6.1  \\ 
\hline
\hline
MOPT~\cite{hurtado20arxiv} & 24.80 & 11.73 & 52.41  & 62.37  & 45.27 & 41.82 & 42.39 & 12.88  & 17.07\\ %
\hline
\hline
Our (single scan) + MOT  & 51.92 & 45.16 & 59.69 & 64.60 & 60.40 & 48.36 & 47.84 & 6.65 & 12.69 \\ 
Our (single scan) + SFP  & 45.45 & 34.61 & 59.69 & 64.60 & 60.40  & 48.24  & 47.72  &  3.01   & 7.93\\ 

Ours (2 scans) & 59.86 & {58.79} & \textbf{60.95} & {64.96} & \textbf{63.06} & {51.14} & {50.67} & \textbf{29.04}  & \textbf{33.2}\\

Ours (4 scans) & \textbf{62.74} & \textbf{65.11} & 60.46 & \textbf{65.36} & 61.26 & \textbf{51.50} & \textbf{51.20} & 0.34    & 4.8 \\

\bottomrule 
\end{tabular}
}
\vspace{-5pt}
\caption{4D Panoptic (validation set). MOT -- \textit{tracking-by-detection} method by~\cite{Weng20iros}; SFP -- \textit{tracking-by-detection} via scene flow based segment propagation~\cite{mittal20cvpr}; PP -- PointPillars~\cite{Lang19CVPR} detector.}
\label{table:baselines4d_val}
\end{table*}
\begin{table*}[h]
\centering
\resizebox{1.0\linewidth}{!}{
\begin{tabular}{lccc|cccc|cc|ccc}
\toprule
Category   & \# Instances & \% Instances & \# Scans & TP    & FP   & FN   & IDS & Precision & Recall & MOTSA &
S$_\text{assoc}$        & S$_\text{cls}$ \\ 
\midrule

\multirow{2}{*}{Motorcycle}
  & \multirow{2}{*}{255}    & \multirow{2}{*}{0.01}   & 2     & 209   & 151  & 46   & 31   & 0.58  & 0.82   & 0.11  & 0.56     & 0.88   \\
 & &   & 4 & 231   & 747  & 24   & 9    & 0.24  & 0.91   & -2.06 & 0.81     & 0.74   \\\midrule

\multirow{2}{*}{Other-vehicle}  &  \multirow{2}{*}{2138}   & \multirow{2}{*}{0.06}  & 2 & 778   & 362  & 1360 & 162  & 0.68  & 0.36   & 0.12  & 0.17     & 0.56   \\
 & & & 4 & 1022  & 1131 & 1116 & 99   & 0.47  & 0.48   & -0.10 & 0.38     & 0.55   \\

\bottomrule
\end{tabular}
}
\vspace{-7pt}
\caption{Per-class evaluation on SemanticKITTI validation set (2 and 4 scan versions).}
\label{table:per_class_cond}
\end{table*}

As can be seen from Tab.~\ref{table:pan_results}, our method achieves state-of-the-art results on all metrics for semantic and panoptic segmentation~\cite{Kirillov19CVPR, Behley19ICCV, Behley20arxiv}. 
The first two entries use two different networks for object detection and semantic segmentation, followed by fusion of the results. We use a single network to obtain both semantic and instance segmentation of the point cloud in a single network pass. We note that the recently proposed Panoptic RangeNet~\cite{milioto2020iros} and RangeNet++~\cite{Milioto19IROS} combined with PointPillars~\cite{Lang19CVPR} detector operate on the range image and not the point cloud, and therefore, use a different backbone. However, KPConv with PointPillars uses the same backbone as our method.

\PAR{4D Panoptic Segmentation.} 
For evaluation in the multi-scan setting for the 4D panoptic segmentation task, we extend all single-scan methods reported in Tab.~\ref{table:pan_results}, except for the Panoptic RangeNet~\cite{milioto2020iros}. %
We adapt them to the sequential domain using two strategies. {AB3DMOT}~\cite{Weng20iros} uses a constant-velocity motion model to obtain track predictions associated with object detection based on a 3D bounding box overlap. 
The second strategy, {Scene Flow Propagation (SFP)} is inspired by standard baselines that perform optical flow warping of points, followed by mask-IoU based association. This approach is commonly used in vision-based video object segmentation~\cite{Luiten18ACCV}, video instance segmentation~\cite{Yang19ICCV}, and multi-object tracking and segmentation~\cite{Osep18ICRA, Osep19arxiv, Voigtlaender19CVPR}. Instead of optical flow, we use state-of-the-art LiDAR scene flow~\cite{mittal20cvpr}. 
We outline our results, obtained on the test set in Tab.~\ref{table:baselines4d}. As can be seen, the baseline that uses {KPConv}~\cite{Thomas19ICCV} to obtain per-pixel classification, {PointPillars (PP) detector}~\cite{Lang19CVPR} and a network for point cloud propagation ({SFP}~\cite{mittal20cvpr}) performs slightly better in terms of association accuracy than the standard 3D MOT baseline. 
Our method that unifies all three aspects in a single network outperforms all tracking-by-detection baselines by a large margin, including our single-scan baseline.%
This confirms the importance of tackling all three aspects of these tasks in a unified manner.  
An important contribution of our paper is the finding that even when processing smaller overlapping sub-sequences with our network (and resolving intra-window associations with a simple overlap-based approach), we perform significantly better compared to single-scan baselines that use more elaborate association techniques (\eg, Kalman filter), as can be confirmed in Tab.~\ref{table:baselines4d}.

\PAR{Metric Insights.} 
In this section, we analyze the performance on the validation split (Tab.~\ref{table:baselines4d_val}) through the lens of several evaluation metrics and analyze per-class performance (Tab.~\ref{table:per_class_cond}). 
Our method outperforms all baselines with respect to all metrics. However, while our 4-scan variant performs better than the 2-scan variant in terms of \metric, we observe a significant drop in the MOTSA score. Our analysis shows that this is caused by negative MOTSA scores on some classes due to a decrease in precision while having fewer ID switches (see Tab.~\ref{table:per_class_cond}, and supplementary).%
We visualize such case in Fig.~\ref{fig:aqvsmotsa_main}. As can be seen, the difference is due to the semantic interpretation of the points and not due to the segmentation and tracking quality at the instance level. This confirms the nonintuitive behavior of MOTSA, while our metric provides insights on both semantic interpretation and instance segmentation and tracking. 
For a more details, we refer to the supplementary.

\section{Conclusion}
In this paper, we extended LiDAR panoptic segmentation to the temporal domain resulting in the 4D Panoptic Segmentation task. We presented an evaluation metric suitable for analyzing this task's performance and proposed a new model. 
Importantly, we have shown that a single model tackling semantic segmentation and point-to-instance association jointly in space and time substantially outperforms methods that independently tackle these aspects. 
We hope that our unified view and model, accompanied by a public benchmark, will pave the road to future developments. 

\small{
\PAR{Acknowledgements.}
This project funded by the Humboldt Foundation through the Sofja Kovalevskaja Award, the EU Horizon 2020 research and innovation programme under grant agreement No.101017008 (Harmony) and the German Federal Ministry of Education and Research (BMBF) under grant No.01IS18036B. The authors of this work take full responsibility for its content. We thank authors of ~\cite{hurtado20arxiv} for providing results for their approach, Ismail Elezi and the whole DVL group for helpful discussions.

}

{\small
\bibliographystyle{ieee_fullname}
\bibliography{abbrev_short,refs}
}

\clearpage
\appendix
\part*{Supplementary Material}
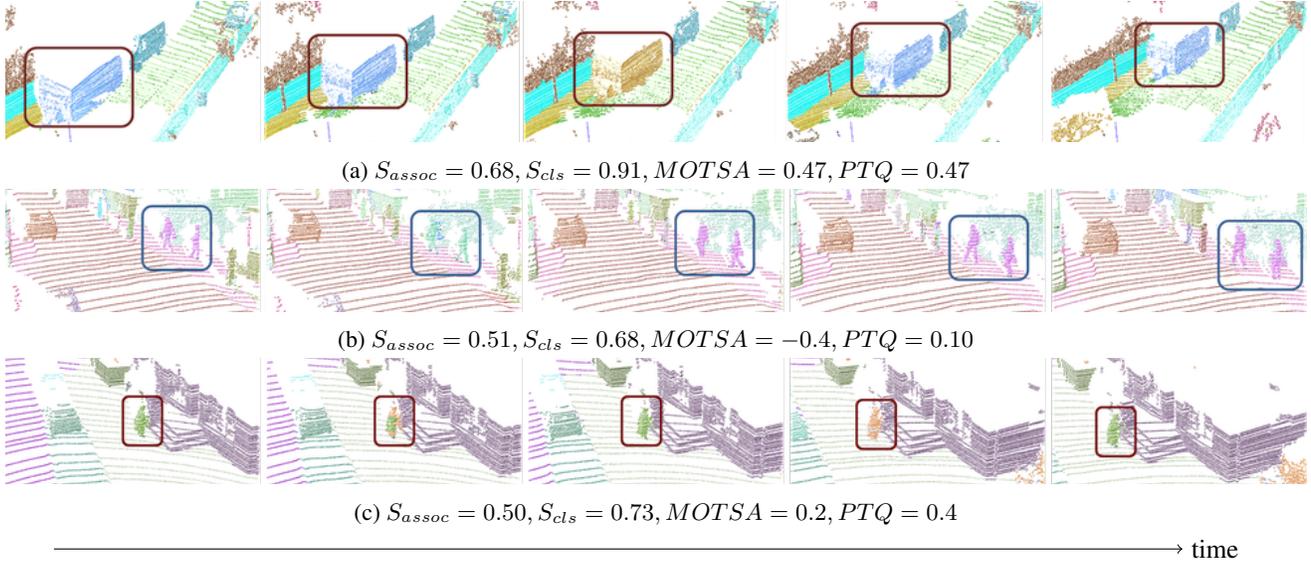
\begin{figure*}[ht!]
     \centering
         \begin{subfigure}{\textwidth}
         \centering
         \includegraphics[trim=0cm 1.0cm 0cm 1.0cm, clip=true, width=0.99\textwidth]{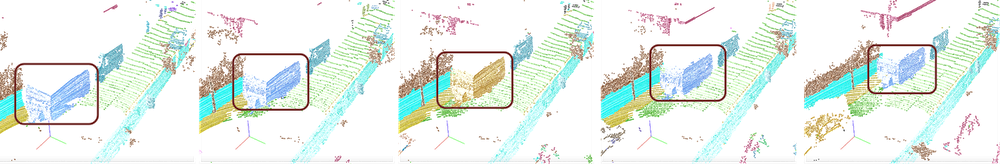}
         \caption{$S_{assoc}=0.68, S_{cls}=0.91, MOTSA=0.47, PTQ=0.47$}
         \label{fig:a}
     \end{subfigure}
     \begin{subfigure}{\textwidth}
         \centering
         \includegraphics[trim=0cm 1.5cm 0cm 1.0cm, clip=true, width=0.99\textwidth]{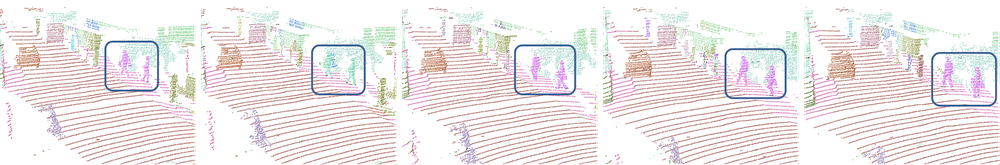}
         \caption{$S_{assoc}=0.51, S_{cls}=0.68, MOTSA=-0.4, PTQ=0.10$}
         \label{fig:b}
     \end{subfigure}
     \begin{subfigure}{\textwidth}
         \centering
         \includegraphics[trim=0cm 1.2cm 0cm 1.2cm, clip=true, width=0.99\textwidth]{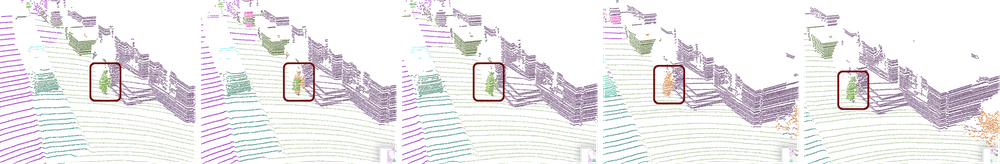}
         \caption{$S_{assoc}=0.50, S_{cls}= 0.73, MOTSA=0.2, PTQ=0.4$}
         \label{fig:c}
     \end{subfigure}
     \begin{tikzpicture}[node distance=2cm]
        \node (A) at (0.25, 0.0) {};
        \node (B) at (15.5, 0.0) {};
        \draw[->, to path={-- (\tikztotarget)}](A) edge (B);
        \node[text width=1cm] at (16.0,0.0){time};
        \end{tikzpicture}
    \caption{Comparison of evaluation metrics for some failure cases. Respective instances which we calculate the metrics are depicted with bounding boxes. In (a) ID recovery is punished by MOTSA and PTQ. In (b) two instances predicted as single instance and in (c) ID switch happened and in the second scan the instance is not segmented correctly.}
    \label{fig:aqvsmotsa}
\end{figure*}

\section{Implementation Details} \label{sec:implementation}
In this section, we (i) provide details about the four different point propagation strategies we experimented with for forming a 4D point clouds and (ii) we detail the point overlap based association procedure we use to link 4D object instances across overlapping point clouds.

\subsection{4D Point Cloud Formation}
Our method works on directly 4D volumes which constructed using consecutive lidar scans. However, due to memory constraints stacking all points is not feasible. To reduce memory usage, when we process the scan $f_{i}$ together with previous scans $f_{i-\tau}$,..., $f_{i-1}$, we take all of the points from $f_{i}$ and sub-sample points from other scans. Moreover, since we already processed previous scans $f_{i-\tau}$,..., $f_{i-1}$ before, we know the semantic class and objectness scores of all points at time step $f$ for that scans. We use three different strategy to sub-sample point from previous scans by leveraging these information.

\PAR{Thing Propagation:} In this strategy, we only sample points from previous scans if the points are assigned to a thing class. If the total number of points are exceeded the gpu memory limit, we randomly sub-sample again. 

\PAR{Importance Sampling:}  We select 10\% of points from a previous scans using the objectness score predicted by the network in the previous time steps. Thus, points with higher objectness scores have a higher chance to be used in the clustering process in the following scans.  

\PAR{Temporal Decay:} In this strategy, we use importance sampling using objectness scores again. However, instead of sampling 10\% of points from each past scan, we select the percentage of points based on temporal proximity of scans. Given a temporal window size of $\tau$, we select the number of points $n_{i}$ as:
\begin{align}
 n_{i} = \frac{e^{i}}{\sum_{n=1}^{\tau-1}e^{i}} , \quad i=1,2,3,\dots,\tau-1,
\end{align}
where $n_{\tau-1}$ is the closest scan to the current scan. In this strategy more points would be sampled from scans which are temporally close.

\PAR{Temporal Stride:} We used importance sampling in this strategy, but instead of using points from previous scans, \ie, \mbox{$i=1,2,3,\dots,\tau-1$}, we used every second scan, \ie, \mbox{$i=1,3,5,\dots,\tau-1$}. For the points from the remaining scans, we assigned predictions by looking at the closest points, which had class and instance prediction. 

\subsection{Clustering}
Our method can cluster points with different semantics and does not provide a single class label for a specific instance. This can be adapted depending on the requirements of the downstream application (\eg, via majority vote). Moreover, if the number of points that assigned to a specific cluster is lower than a threshold, we eliminate that instance from the final prediction.

\subsection{Tracking}

As discussed in the main paper (Section 3), we process multiple scans together in an overlapping fashion. 
For a window size of $\tau$, at time  $t$, we process scans $f_{i-\tau}^{t}, \dots, f_{i}^{t}$ together by overlapping them in a 4D point cloud. $f_{i}^{t}$ represent the scan $i$ which processed at time step $t$. 

To associate instances at time $t$ and $t+1$, we look at instance intersections in scans which are common in both time steps. For instance, with temporal window size of two, we would process scans $f_{1}^{1}$ and $f_{2}^{1}$, next we would process $f_{2}^{2}$ and $f_{3}^{2}$ together. To transfer ids from the previous time to the current scan ($f_{3}^{2}$), we would look the instance intersections in scans which processed on both time step ($f_{2}^{1}$ and $f_{2}^{2}$). Since the instance ids are same for the scans which processed together ( $f_{2}^{2}$ and $f_{3}^{2}$), the association would be finished between overlapping 4D volumes. 

For the intersection, we consider all common scans. When there is a conflict (i.e, one instance has overlap with two instance in the next step), we pick the instance pair which have higher intersection-over-union. If any of the intersections do not surpass IoU of $0.5$, we create a new ID for the instance.

\section{Additional Results}

\subsection{Ablation on the Temporal Window Size}

In Tab.~\ref{table:abl_nframes}, we highlight the performance of our method for temporal window size $\tau = {1, 2, 3, 4, 6, 8}$. As can be seen, the association accuracy is increasing up to $\tau=4$ and then saturates, while classification accuracy saturates a $\tau=2$; however, it only decreases marginally.

\begin{table}[t]
\footnotesize
\centering
\resizebox{1.0\linewidth}{!}{
\begin{tabular}{c|ccc|cc}
\toprule
\# Scans          & \metric           & S$_\text{assoc}$ & S$_\text{cls}$       &  IoU$^\text{St}$      & IoU$^\text{Th}$            \\ \midrule

1 & 51.92           & 45.16             & 59.69             & 64.60             & 60.40    \\
2 & 59.86           & 58.79             & \textbf{60.95}    & 64.96             & \textbf{63.06}  \\
3 & 61.74           & 62.65             & 60.85             & 65.16             & 62.53  \\
4 & \textbf{62}.74  & \textbf{65.11}    & 60.46             & \textbf{65.36}    & 61.26  \\
6 & 61.52           & 64.28             & 58.88             & 65.32             & 57.38  \\
8 & 59.09           & 62.30             & 57.68             & 65.23             & 54.52  \\
\bottomrule
\end{tabular}
}
\vspace{-5pt}
\caption{Panoptic Tracking on SemanticKITTI valid. set.}
\label{table:abl_nframes}
\end{table}

\subsection{Per-class Evaluation} \label{sec:per_class}

In this section, we analyze the performance on the validation split (Tab.~\ref{table:baselines4d_val}) through the lens of several evaluation metrics and analyze per-class performance in Tab.~\ref{table:per_class_full} (this table extends Tab.~\ref{table:per_class_cond} from the main paper). 
While our 4-scan variant performs better than the 2-scan variant in terms of \metric, we observe a significant drop in the MOTSA score. 
Our analysis shows that this is because we obtain negative MOTSA scores on some classes due to a decrease in precision while having fewer ID switches. This unintuitive behavior of MOTSA can be further validated when analysing performance for class, \eg, \textit{other-vehicle}. For this class the IDS reduces ($162 \to 99$), the precision drops ($0.68 \to 0.47$), while recall improves from ($0.36 \to 0.47$). 
In our metric, this is reflected in the decrease of $\sseg$ ($0.56 \to 0.55$) and increase in $\sassoc$ ($0.17 \to 0.38$) while MOTSA unintuitively drops from $0.12$ to $-0.1$, even though association capabilities improve.  %

We visualize such cases in Fig.~\ref{fig:aqvsmotsa}. As can be seen, the difference is due to the semantic interpretation of the points and not due to the segmentation and tracking quality at the instance level. This confirms the nonintuitive behavior of MOTSA, while our metric provides insights on both semantic interpretation and instance segmentation and tracking. 
As shown in Figure \ref{fig:a}-\ref{fig:c}, our method successfully recovers the ID of the instance. This behavior is penalized by both MOTSA and PTQ, but not by the association score of our metric $S_{assoc}$. Moreover, while the instances tracked reasonably well in Figure~\ref{fig:b}, MOTSA and PTQ scores decrease substantially due to poor segmentation of the instances.

Finally, we acknowledge that our method works very well on the most frequently occurring object classes (\textit{car}), however, segmenting and tracking objects that appear in the long tail of the object class distribution remains challenging.

\begin{table*}[h]
\footnotesize
\centering
\resizebox{1.0\linewidth}{!}{
\begin{tabular}{llll|cccc|ccc|cc}
\toprule
Category  & \# Scans & \# Instances & \% Instances  & TP    & FP   & FN   & IDS & Prec. & Recall & MOTSA & S$_\text{assoc}$        & S$_\text{cls}$ \\
\midrule
Car     & 2    & 29255  & 0.80   & 27553 & 687  & 1702 & 1204 & 0.98  & 0.94   & 0.88  & 0.72     & 0.96   \\
    & 4       &  &    & 27401 & 845  & 1854 & 720  & 0.97  & 0.94   & 0.88  & 0.77     & 0.96   \\\midrule 
    
Truck   & 2      & 1253   & 0.03   & 447   & 226  & 806  & 90   & 0.66  & 0.36   & 0.10  & 0.15     & 0.38   \\
     & 4    &    &    & 496   & 331  & 757  & 52   & 0.60  & 0.40   & 0.09  & 0.20     & 0.39   \\\midrule
     
Bicycle   & 2   & 792    & 0.02   & 435   & 132  & 357  & 64   & 0.77  & 0.55   & 0.30  & 0.36     & 0.72   \\
      & 4 &     &    & 574   & 230  & 218  & 43   & 0.71  & 0.72   & 0.38  & 0.59     & 0.71   \\\midrule
      
Motorcycle & 2    & 255    & 0.01   & 209   & 151  & 46   & 31   & 0.58  & 0.82   & 0.11  & 0.56     & 0.88   \\
   & 4 &     &    & 231   & 747  & 24   & 9    & 0.24  & 0.91   & -2.06 & 0.81     & 0.74   \\\midrule

Other-vehicle & 2 & 2138   & 0.06   & 778   & 362  & 1360 & 162  & 0.68  & 0.36   & 0.12  & 0.17     & 0.56   \\
  & 4 &    &   & 1022  & 1131 & 1116 & 99   & 0.47  & 0.48   & -0.10 & 0.38     & 0.55   \\\midrule

Person & 2       & 1975   & 0.05   & 1183  & 282  & 792  & 203  & 0.81  & 0.60   & 0.35  & 0.31     & 0.65   \\
        & 4 &    &    & 1180  & 346  & 795  & 143  & 0.77  & 0.60   & 0.35  & 0.35     & 0.63   \\\midrule

Bicyclist & 2     & 816    & 0.02   & 720   & 39   & 96   & 33   & 0.95  & 0.88   & 0.79  & 0.63     & 0.89   \\
      & 4 &     &    & 750   & 39   & 66   & 28   & 0.95  & 0.92   & 0.84  & 0.69     & 0.91   \\\midrule

Motorcyclist & 2 & 78     & 0.01   & 0     & 0    & 78   & 0    & 0.00  & 0.00   & 0.00  & 0.10     & 0.00   \\
  & 4 &      &   & 0     & 0    & 78   & 0    & 0.00  & 0.00   & 0.00  & 0.16     & 0.00  \\ \bottomrule

\end{tabular}
}
\caption{Per-class tracking evaluation on Semantic-KITTI validation set (2 and 4 scan versions).}
\label{table:per_class_full}
\end{table*}

\end{document}